\documentclass{article}

\usepackage[final]{neurips_2024}

\usepackage[utf8]{inputenc} 
\usepackage[T1]{fontenc}    
\usepackage{hyperref}       
\usepackage{url}            
\usepackage{booktabs}       
\usepackage{amsfonts}       
\usepackage{nicefrac}       
\usepackage{microtype}      
\usepackage{xcolor}         
\usepackage{graphicx}
\usepackage{subcaption}

\usepackage[ruled,vlined]{algorithm2e}
\usepackage{amsmath}

\usepackage{floatrow}

\usepackage[backgroundcolor=gray!15]{mdframed}

\title{Thinking Fast and Laterally: Multi-Agentic Approach for Reasoning about Uncertain Emerging Events}

\newcommand{\ourSystem}{\textsc{SaLT}}

\author{%
  Stefan Dernbach \\
  \texttt{stefan.dernbach@pnnl.gov} \\
  \And
  Alejandro Michel \\
  \texttt{alejandro.michelzuniga@pnnl.gov} \\
  \And
  Khushbu Agarwal \\
  \texttt{khushbu.agarwal@pnnl.gov} \\
  \And
  Christopher Brissette \\
  \texttt{cbrissette@nvidia.com} \\
  \And
  Geetika Gupta \\
  \texttt{gegupta@nvidia.com} \\
  \And
  Sutanay Choudhury \\
  \texttt{sutanay.choudhury@pnnl.gov} \\
}

\begin{document}

\maketitle

\begin{abstract}

This paper introduces lateral thinking to implement System-2 reasoning capabilities in AI systems, focusing on anticipatory and causal reasoning under uncertainty. We present a framework for systematic generation and modeling of lateral thinking queries and evaluation datasets. We introduce Streaming agentic Lateral Thinking ($\ourSystem$), a multi-agent framework designed to process complex, low-specificity queries in streaming data environments. $\ourSystem$ implements lateral thinking-inspired System-2 reasoning through a dynamic communication structure between specialized agents. Our key insight is that lateral information flow across long-distance agent interactions, combined with fine-grained belief management, yields richer information contexts and enhanced reasoning. Preliminary quantitative and qualitative evaluations indicate $\ourSystem$'s potential to outperform single-agent systems in handling complex lateral reasoning tasks in a streaming environment.

\end{abstract}
\section{Introduction}

The rapid advancement of artificial intelligence (AI) has revolutionized our ability to process and understand vast amounts of text and multi-modal content. State-of-the-art AI models now demonstrate remarkable capabilities in answering complex hypothetical questions based on user-provided context, exhibiting traits reminiscent of "System 1" thinking as described by \cite{kahneman2011thinking}.

\begin{figure}[htb]\centering
    {\includegraphics[width=0.9\textwidth]{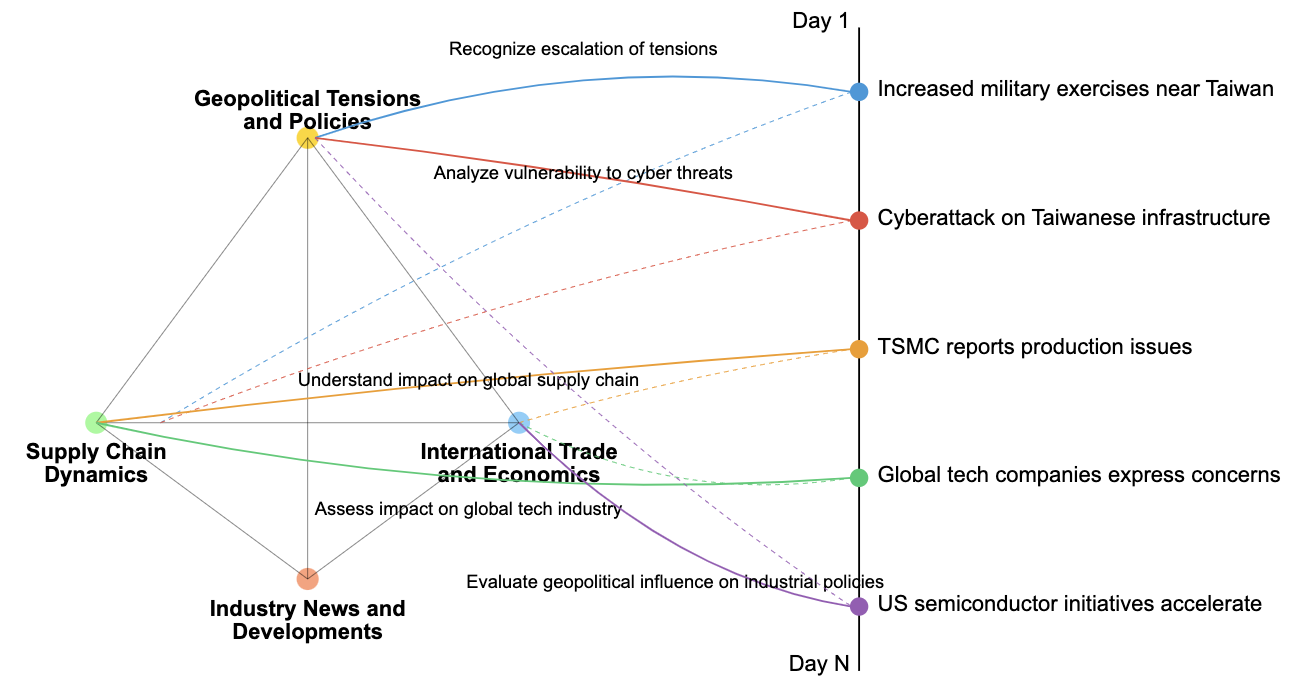}}
\caption{Conceptual overview of the $\ourSystem$ framework. This diagram illustrates the multi-agent structure and dynamic communication flow in processing lateral reasoning queries, such as "monitor the risk to the supply chain of American semiconductor companies". The right side shows a timeline of events that gradually unfold in a datastream over time. The left side depicts the specialized agent network, representing different topics in the problem space. This network structure is dynamically generated to address the specific lateral reasoning query. Connecting lines between agents and events indicate the reasoning criteria employed by each agent to interpret the temporal context.}
\label{fig:architecture}
\end{figure}

A key component of human intelligence is the ability to reason under uncertainty and anticipate future events. Many critical events in personal, financial, and security domains do not occur instantaneously but are preceded by identifiable precursors. For instance, a user might be interested in monitoring risks to a specific industry supply chain arising from emerging geopolitical events (Fig. \ref{fig:architecture}). Recognizing early signals and tracking their development over time enables proactive planning and appropriate responses as events unfold.

These cognitive abilities require sophisticated reasoning about: 1) possible event chains under uncertainty, 2) dynamic evaluation of event chain probabilities as new information emerges, and 3) intelligent aging and retrieval of information from long-term to short-term memory. Such traits are hallmarks of "System 2" thinking, essential for advanced cognitive processes that distinguish human-level intelligence (\cite{kahneman2011thinking}).

Lateral thinking, introduced by Edward de Bono in the late 1960s (\cite{de1970lateral}), involves solving problems through indirect, creative reasoning that may not be immediately obvious. This approach is particularly valuable in complex, dynamic environments where conventional methods fall short. We posit that a multi-agent architecture \cite{wu2023autogen} presents a promising solution for continuously processing a large volume of anticipatory questions. By leveraging specialized agents for distinct domains, this system enables parallel processing of domain-specific data and facilitates cross-domain connections. This distributed approach mirrors the non-linear pathways of lateral thinking, allowing for the discovery of novel patterns that traditional, sequential reasoning methods would miss.

In this work, we introduce a systematic framework for generating and modeling lateral thinking queries and associated evaluation datasets. This framework addresses the challenge of low-specificity queries in real-world applications, where individuals can express interest in specific outcomes but cannot describe every possible combination of event sequences leading to those outcomes. Building on this framework, we present Streaming agentic Lateral Thinking ($\ourSystem$), a multi-agent system designed to simultaneously process \textsl{multiple} complex, low-specificity queries in streaming data environments. $\ourSystem$ implements lateral thinking-inspired System-2 reasoning through a dynamic communication structure between specialized agents.

$\ourSystem$ operates by monitoring evolving information around assigned topics, with agents exchanging probabilistic observations (subsequently referred to as ``belief statements") to form and temporally maintain hypotheses about query-focused outcomes. Unlike static multi-agent systems, $\ourSystem$'s agent network's communication structure continuously adapts based on the relevance of agent expertise and the newly formed beliefs. This dynamic topology enables the system to find unconventional pathways for information flow, addressing scenarios where information is continuously changing and the immediate context rarely contains the full answer to a user query.

In summary, we make the following contributions:

\begin{enumerate}
    \item A systematic framework for generating and modeling lateral thinking queries and evaluation datasets, providing a foundation for research in anticipatory reasoning under uncertainty.
    \item Streaming agentic Lateral Thinking ($\ourSystem$): A novel multi-agent architecture (Fig. \ref{fig:architecture}) that implements lateral thinking-inspired System-2 reasoning for simultaneous processing of multiple anticipatory queries in streaming data environments.
\end{enumerate}

\section{Related Work}

From an algorithmic perspective, a complex event can be decomposed into sub-events.  See Fig. \ref{fig:salt_arctic} and \ref{fig:salt_africa} for more illustrative examples. The process of robust prediction of such complex events can generally be reduced to three steps: 1) accurate identification of its sub-events when or before they occur, 2) the ability to connect these sub-events as they emerge over time through a single or multiple data sources, and 3) maintenance of the combined or partially materialized event information such that that memory complexity does not increase monotonically and lead to system failure.  Each of these steps has been extensively studied across various fields in computer science, including databases, natural language processing, and symbolic or probabilistic reasoning methods. Drawing strong inspiration from this prior literature, $\ourSystem$ introduces large-language model \cite{achiam2023gpt, ram2023context}-based implementation of each of these steps.

\textbf{Streaming query/reasoning} Complex event processing has been studied extensively over the past three decades, initially motivated by the advent of online financial markets and subsequently by the growth of cyber data streams and social media. A common approach among these prior solutions \cite{chandrasekaran2003telegraphcq, choudhury2015selectivity, della2009s}: 1) A query decomposition approach where the complex query processing task is decomposed into a data-flow graph, with nodes representing operators that process a stream through SQL, graph, or different models, and 2) an event graph that guides how partial results are joined to produce a final answer.  Maintenance of materialized partial query results and their aging over time is a key performance issue for such systems  \cite{barbieri2010incremental, ren2011optimising}.  As discussed in the next section, $\ourSystem$ derives these decomposition graphs from user-specified queries, initiates an agent network structure \cite{wu2023autogen}, and pursues a dynamic inter-agent communication strategy for dataflow.

\textbf{Induction and exploitation of Knowledge Schema} Scaling up such methods on unstructured, noisy data streams extracted from text and other modalities introduces new challenges, including query event interpretation from text \cite{lee2015event}, induction of unseen events \cite{huang2020semi}, and evolution of underlying knowledge schema \cite{yu2021lifelong}.  In the current work, we do not restrict $\ourSystem$ to any specific event or knowledge (graph) schema.  This choice is primarily driven by the goal to support multiple user queries drawn from a wide range of diverse but related domains. Recent work \cite{ghafarollahi2024sciagents} illustrates the benefits of grounding the beliefs generated in an LLM-Agent network via a knowledge graph.

\textbf{Probabilistic approach} Probabilistic approaches to event detection have been studied in the past for structured data streams, with ``possible worlds" representing a probabilistic assignment of various events occurring in the data stream \cite{cormode2007sketching}. Formulation of query processing operators in a Hidden Markov Model fashion and development of probabilistic stream algebra were proposed by \cite{re2008event}.  In the broader context beyond event detection, belief propagation methods have been the cornerstone of probabilistic graphical models \cite{murphy2013loopy}.  More recently, loopy belief propagation methods have been combined with graph neural networks to propose neural message propagation with embeddings \cite{satorras2021neural}, inspiring the implementation of the message propagation and belief synthesis steps in our proposed algorithm. 

\section{Methods}

Streaming agentic Lateral Thinking ($\ourSystem$) is designed to process streaming information based on a set of user-registered questions, continuously generating hypotheses about query matches by synthesizing the evolving reasoning outputs produced by multiple specialized agents. The key components and mechanisms of $\ourSystem$ are as follows:


\subsection{System Architecture}

The components of the $\ourSystem$ architecture are defined as follows.

\begin{itemize}
    \item \textbf{Agents}: Each agent $A_i$ in the network is associated with a set of topics $T_i = \{t_1, t_2, ..., t_n\}$. See Fig. 1 and the appendix for examples.  In a multi-query setting, we compute an union of all query specific topics and use the resultant set to initilize the network.
    \item \textbf{Network Topology}: The agents are connected in a graph structure $G(V, E)$, where vertices $V$ represent agents and edges $E$ represent information flow pathways between agents. This graph dynamically evolves based on the context of each batch in the stream.
    \item \textbf{Belief Statements}: Agents generate and maintain a set of belief statements $B_i = \{b_1, b_2, ..., b_m\}$. Each belief statement $b_j$ is characterized by a tuple $(s, c, t, r)$, where $s$ is the statement text, $c$ is a confidence score, $t$ is a timestamp, and $r$ is a list of relevant reference ids.  We provide an example of a belief statement in the next page.
    
\end{itemize}

\textsc{Problem definition}  Given the aforementioned setup, $\ourSystem$ aims to optimize the function $f(S, Q, A, G) \rightarrow H$, where $S = \{s_1, s_2, ..., s_t, ...\}$ is an incoming data stream, $Q = \{q_1, q_2, ..., q_m\}$ is a set of user queries, $A = \{A_1, A_2, ..., A_n\}$ is a set of specialized agents, and $G(V, E)$ is a dynamic graph representing inter-agent connections. 

For each batch $s_t$, $\ourSystem$ updates $G$, generates and refines belief statements $B_i(t)$ for each agent $A_i$, and synthesizes these to produce a set of hypotheses $H(t) = \{h_1, h_2, ..., h_k\}$. The objective is to maximize $\sum_i \text{relevance}(h_i, Q)$ and $\sum_i \text{accuracy}(h_i)$ while minimizing computational complexity $C(f)$ and ensuring $|B_i(t)| \leq M$ for some memory bound $M$, $\forall i, t$. This formulation encapsulates $\ourSystem$'s goal of continuously generating accurate, relevant hypotheses from streaming data using a dynamically adapting multi-agent system, subject to real-time processing and memory constraints.  

\begin{mdframed}
    \textbf{Question:} What actions are American semiconductor companies likely to take in response to these vulnerabilities and risks?\\
    \textbf{Context}: [Query context can be initialized from a batch of articles or prior belief statements.] \\
    \textbf{Response:} American semiconductor companies are likely to take several actions, including diversifying their supply chains to reduce dependency on any single region, investing in increased cybersecurity measures to protect against cyberattacks, and advocating for diplomatic solutions to reduce geopolitical tensions. They might also increase collaboration with international partners to bolster their supply chain resilience and seek governmental support for policies that encourage domestic manufacturing capabilities.\\
    \textbf{Confidence:} 0.9\\
    \textbf{Timestamp:} 127\\
    \textbf{References:} [article-103, article-104]
\end{mdframed}

We now proceed to describe three key steps in $\ourSystem$: initialization of the agent network, initiation of new belief statements and updating pre-computed beliefs, and sharing across the network topology to synthesize new belief statements. The latter two steps are executed iteratively for each batch of incoming streaming data.

\textbf{Agent Initialization} $\ourSystem$ is first initialized by translating the user-specified question set into a set of topics. In Algorithm~\ref{alg:initializenetwork}, the \texttt{ExtractTopics} function uses a pre-trained topic classification model to identify relevant topics from the input questions. The \texttt{CreateAgent} function handles the construction of simple RAG agents, and the \texttt{BuildGraph} function creates initial connections between agents based on topic similarity. An example output from \texttt{ExtractTopics} is provided in section \ref{subsec:Agent network initialization}.

\begin{algorithm}[H]
\SetAlgoLined
\KwIn{Set of questions $Q$}
\KwOut{Set of agents $A$, Graph $G$}
$T \gets \text{ExtractTopics}(Q)$\;
$A \gets \{\}$\;
\ForEach{unique topic $t_i$ in $T$}{
    $A_i \gets \text{CreateAgent}(t_i)$\;
    $A \gets A \cup \{A_i\}$\;
}
$G \gets \text{BuildGraph}(A)$\;
\Return{$(A, G)$}\;
\label{alg:initializenetwork}
\caption{InitializeNetwork}
\end{algorithm}

\textbf{Stream Processing} $\ourSystem$ processes streaming information in multiple steps. First, the input query (that can be an article, image or a batch of them) assigns a relevance score relative to each expert agent, and are processed only by agents with sufficiently high relevance. We use a cosine similarity between the query and agent representation. Each agent constructs new beliefs based on assigned articles  Algorithm~\ref{alg:processinformation}.

\textbf{Belief Propagation and Synthesis} Belief propagation and synthesis are key to $\ourSystem$'s emergent lateral thinking capabilities. Algorithm~\ref{alg:propagatebeliefs} shows how each agent within $\ourSystem$ takes in the most relevant beliefs from its neighbors, and alters its own beliefs based on this information. 

\begin{algorithm}[H]
\SetAlgoLined
\KwIn{Information stream $I$, Set of agents $A$, Graph $G$}
\ForEach{information item $i$ in $I$}{
    $R \gets \text{AssignRelevance}(i, A)$\;
    \ForEach{agent $A_j$ in $A$ where $R_j > \text{threshold}$}{
        $B_{new} \gets A_j.\text{ProcessItem}(i)$\;
        $A_j.B \gets A_j.B \cup B_{new}$\;
        $\text{PropagateBeliefs}(A_j, B_{new}, G)$\;
    }
    $G \gets \text{UpdateTopology}(G, R)$\;
}
\label{alg:processinformation}
\caption{ProcessInformation}
\end{algorithm}

\begin{algorithm}[H]
\SetAlgoLined
\KwIn{Source agent $A_s$, New beliefs $B_{new}$, Graph $G$}
\ForEach{agent $A_r$ connected to $A_s$ in $G$}{
    $R \gets \text{ComputeRelevance}(A_s, A_r, B_{new})$\;
    \If{$R > \text{threshold}$}{
        $B_{shared} \gets A_s.\text{ShareBeliefs}(A_r, B_{new})$\;
        $B_{synthesized} \gets A_r.\text{SynthesizeBeliefs}(B_{shared})$\;
        $A_r.B \gets A_r.B \cup B_{synthesized}$\;
    }
}
\label{alg:propagatebeliefs}
\caption{PropagateBeliefs}
\end{algorithm}

\subsection{Dynamic Network Topology}

While the multi-agent system described in prior subsections has a defined graph structure, all connected agents do not necessarily communicate with each other for every query. Consider a system which has two agents, one that specializes in the ``weather events", and another specializes in ``agricultural commodities market". Assume these two agents are connected via the network, and the purpose of such a multi-agent system is to predict how extreme weather events may affect the farmimg industry in the Pacific Northwest. Given a set of articles that pertains only to weather, it does not necessarily make sense to share those belief statements with our ``commodities" agent. 

Due to instances like the above, $\ourSystem$'s network topology evolves based on the relevance and of agent connections as seen in Equation~\ref{eq:agent_conn} which follows.

\begin{equation}
w_{ij} = \alpha \cdot \text{sim}(T_i, T_j)
\label{eq:agent_conn}
\end{equation}

Here $w_{ij}$ is the connection weight between agents $A_i$ and $A_j$, which handle topics $T_i$ and $T_j$ respectively. $\text{sim}(T_i, T_j)$ is the topic similarity and $\alpha$ is a coefficient which controls the relative importance of this particular connection. This construction is important for avoiding simple diffusion-like dynamics. If there were only a communication network structure without any further selective behavior, the network of agents will eventually converge on a nearly static belief set.
\section{Dataset}
To rigorously evaluate $\ourSystem$'s performance, we developed a comprehensive dataset of lateral thinking questions and scenarios, accompanied by novel complexity metrics. This section details our data generation process and benchmarking methodology.

\textbf{Dataset Creation} We generated a synthetic dataset comprising 30 lateral thinking queries using a combination of state-of-the-art language models (Claude-3.5 and GPT-4o) and human expert curation. Each query was crafted to investigate trends with potential cross-industry impacts such as ``Monitor extreme weather patterns for impacts on agricultural commodity prices". We also generated multiple scenarios for each query, designed to emulate real-world situations typically reported in mainstream news media.  For example, the ``Monitor extreme weather patterns for impacts" query can be mapped to three scenarios: a) ``Severe droughts in Brazil lead to surge in coffee futures prices", b) ``Unexpected frost in Florida causes orange juice futures to spike" and c) ``Extended heatwave in Europe results in wheat shortage and price increase".

\textbf{Article Generation} For each scenario, we generated between 20 and 100 articles, each presenting unique developments related to the specific scenario. To ensure diversity and realism in the corpus, we parameterized the article generation process along several dimensions: a) perspective (e.g., conservative, liberal), b) length (ranging from 200 to 1000 words), c) temporal distribution (e.g., uniformly distributed over a 6-month period vs. clustered within a 2-week window).  

\subsection{Lateral Reasoning Complexity Metrics}

To quantitatively model the workload complexity and ensure a balanced distribution, we propose four novel metrics for each query-scenario pair. We'll illustrate each metric using our Taiwan Strait tensions scenario from Fig. 1:

\begin{enumerate}
    \item \textbf{Lateral Measure ($\mathcal{L}$):} Quantifies the number of intermediate causal steps between the trigger event and the outcome. Range: 1-7.
    \textit{Example:} $\mathcal{L} = 3$ for our Taiwan scenario: increased military presence $\rightarrow$ disruption of global semiconductor supply chains $\rightarrow$ stock market volatility.


    \item \textbf{Time Lag Complexity ($\mathcal{T}$):} Measures the temporal distance between the trigger event and its final impact, in days. Range: 0-21 (for this rapid-evolving scenario).  Example: $\mathcal{T} = 14$ for our Taiwan scenario, where significant economic impacts manifest two weeks after the initial spike in tensions.

    \item \textbf{Outcome Uncertainty Coefficient ($\mathcal{U}$):} Quantifies the unpredictability of the relationship between the trigger and its effects on a normalized scale. Range: 0-1.
    \textit{Example:} $\mathcal{U} = 0.9$ for our Taiwan scenario, due to the highly volatile nature of geopolitical crises and their potential for rapid escalation or de-escalation.
\end{enumerate}

\begin{figure}[ht]
    \centering
    \begin{subfigure}{0.5\textwidth}
        \centering
        \includegraphics[width=\textwidth]{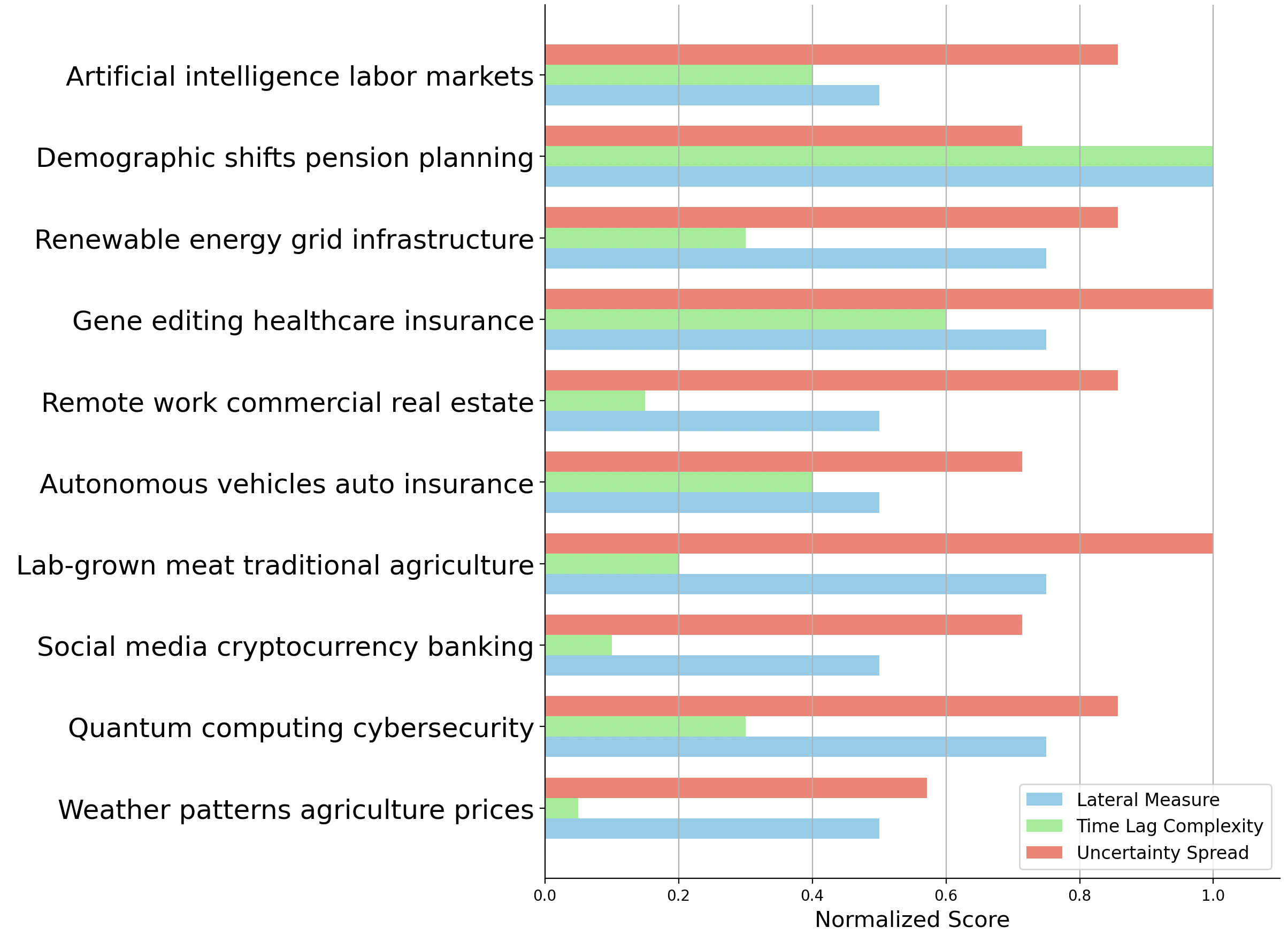}
        \caption{Query Complexity Distribution}
        \label{fig:query_metrics_dist}
    \end{subfigure}
    \hfill
    \begin{subfigure}{0.45\textwidth}
        \centering
        \includegraphics[width=\textwidth]{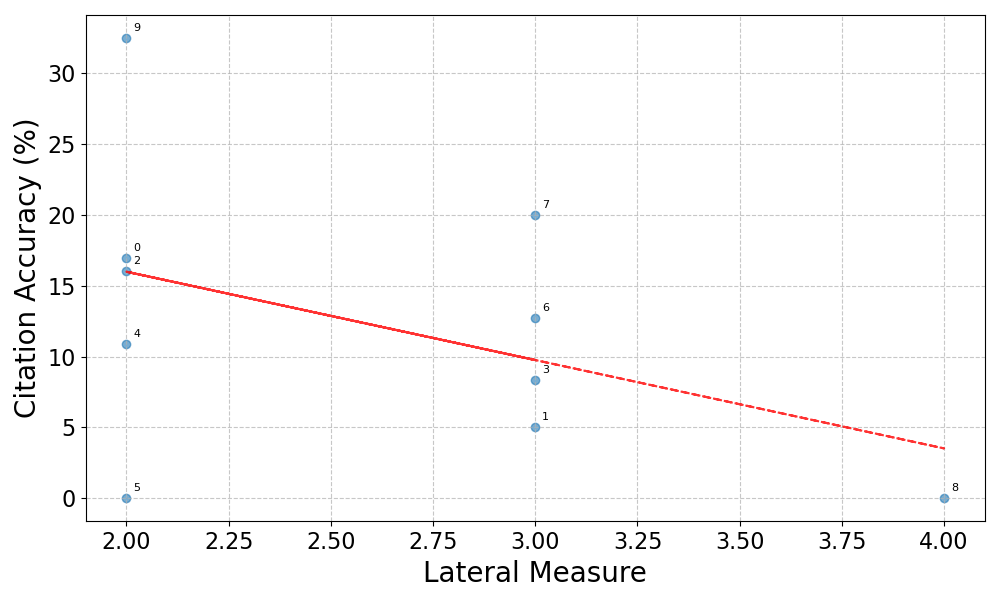}
        \caption{Reasoning accuracy drops for queries with increased lateral measure}
        \label{fig:accuracy_vs_lateral}
    \end{subfigure}
    \caption{(a) Shows normalized lateral measure, time lag complexity, and uncertainty spread across various anticipatory queries. (b) Shows an evaluate of $\ourSystem$ over a dataset with high time-lag complexity and profiles performance as a function of query complexity.}
    \label{fig:query_complexity}
\end{figure}

\textbf{Diversity Assurance}: To ensure a wide range of lateral thinking challenges, we categorized queries into 10 broad domains (e.g., geopolitics, technology, climate change) and ensured balanced representation across these categories (Fig. \ref{fig:lt_queries}). We also varied the complexity levels (based on our proposed metrics) across the dataset.  Fig. \ref{fig:query_metrics_dist} illustrates the diversity of the queries via complexity metrics discussed above.  For illustration of a single query and it's metrics see Fig. \ref{fig:lt_query_metrics} in appendix.



\section{Experiments}

In this preliminary study, we focus on two primary research questions:

\begin{enumerate}
    \item \textbf{RQ1:} How does multi-agentic lateral reasoning compare with single agent approaches?
    \item \textbf{RQ2:} How does the system performance change as a function of query complexity?
\end{enumerate}

\subsection{Experimental Setup}

We explore four model variants in our experiments as listed below.  As the name suggests, SingleTemporalReasoningAgent is an agent that processes the batched data stream and executes as prompt on a rolling window of historical articles.  It runs each of the user-specified queries separately with no information exchange between multiple query execution. See section \ref{sec:Baseline Implementations} for implementation details.

\begin{itemize}
    \item \textbf{Temporal-Claude:} SingleTemporalReasoningAgent with Claude-3.5-Sonnet
    \item \textbf{Temporal-GPT4o:} SingleTemporalReasoningAgent with GPT-4o
    \item \textbf{$\ourSystem$-Claude:} $\ourSystem$ with Claude-3.5-Sonnet
    \item \textbf{$\ourSystem$-GPT4o:} $\ourSystem$ with GPT-4o
\end{itemize}

For brevity, we refer to SingleTemporalReasoningAgent with Claude (or GPT-4o) as Temporal-Claude (or Temporal-GPT4o) in subsequent discussions. We use our custom dataset of lateral thinking queries, as described in Section 3. For this preliminary study, we selected a subset of 30 queries from the full dataset to evaluate our models.  This relatively small set was chosen to ensure we can manually verify the quality of generated articles and reasoning output from each target model.

\subsection{Evaluation Metrics}

We propose two metrics to capture key aspects of lateral thinking:

\textbf{Retrieval Performance} RP measures the system's ability to identify relevant information from diverse sources. A high RP indicates effective navigation through a large corpus, pinpointing pertinent articles even when connections are not immediately obvious—a crucial requirement for lateral thinking.

\begin{equation}
    \text{RP} = \frac{\text{Number of relevant article IDs cited}}{\text{Total number of relevant articles}} \times 100\%
\end{equation}

\textbf{Hypothesis Quality} HQ assesses the system's capacity to synthesize information into meaningful hypotheses. It evaluates the ability to connect disparate pieces of information, forming information rich and specific conclusions.

\begin{equation}
    \text{HQ} = \frac{\text{Number of correctly identified sub-events}}{\text{Total number of sub-events}} \times 100\%
\end{equation}

\subsection{Results}

\textbf{Lateral context augmentation drives outperformance} Results comparing $\ourSystem$  and SingleTemporalAgents are provided in Table \ref{tab:performance-comparison}. We observe that our agent network-based system achieves significantly better performance across both hypothesis generation and citation retrieval compared to the single agent based systems. $\ourSystem$-Claude-Sonnet-3.5  outperforms its corresponding baseline implementation by 39.77\% and 29\% in citation retrieval and hypothesis quality. Salt-GPT-4o outperforms Temporal-GPT-4o by 60\% and 87.63\%. Overall, Claude-Sonnet-3.5 based agents perform better than GPT-4o in retaining facts and generating overall hypotheses.

\textbf{Key insight} While our study is preliminary and larger-scale experiments are needed to conclusively establish the superiority of lateral thinking, a consistent pattern emerges: $\ourSystem$'s reasoning is driven by a significantly larger and diverse context size as measured by the number of belief statements and supporting article count. This suggests that the key to it's success lies in its ability to perform lateral thinking over related contexts, sometimes multiple hops away in the agent network. The system's capacity to ensure that relevant information from distant articles is communicated to the appropriate agents appears to be crucial. However, further research is necessary to fully understand the information flow within the agent network and to validate this hypothesis. This insight, if confirmed, could represent a significant advancement in an AI system's ability to reason across diverse and seemingly unrelated information sources arriving at different points of time.

\textbf{Qualitative Analysis}  Answers generated by $\ourSystem$ are more a) information-rich and b) characterized by higher specificity.  Below we present a brief case study analyzing results to the query: ``Monitor if any new economic activity develops as a result of changing climate patterns".

\begin{enumerate}
\item Comprehensiveness of assessment:
   $\ourSystem$-GPT-4o and $\ourSystem$-Claude provide more comprehensive assessments covering a broader range of economic activities. For example:
   \begin{itemize}
   \item $\ourSystem$-GPT-4o mentions "innovations in maritime technologies, including advanced ice-breaking capabilities and improved navigation systems" and "growing economic activity in climate-adaptive technologies across various sectors."
   \item $\ourSystem$-Claude discusses "innovation in sustainable technologies, including renewable energy, water management systems, and climate-resilient agriculture."
   \end{itemize}
   In contrast, Temporal-GPT-4o and Temporal-Claude focus primarily on Arctic shipping routes and resource exploration.

\item Depth of analysis on long-term implications:
   $\ourSystem$ versions offer deeper insights into long-term economic consequences:
   \begin{itemize}
   \item $\ourSystem$-GPT-4o states: "These developments are reshaping global trade patterns and potentially shifting economic power towards nations with strong Arctic presences or advanced maritime technologies."
   \item $\ourSystem$-Claude notes: "Overall, while climate change is creating new economic opportunities, it is also fundamentally altering global power dynamics, security concerns, and governance structures, which will have long-term implications for international trade and economic partnerships."
   \end{itemize}
   The Temporal-GPT-4o and Temporal-Claude outputs do not address these broader implications.

\item Balanced perspective on opportunities and challenges:
   $\ourSystem$ outputs provide a more nuanced view of both opportunities and risks:
   \begin{itemize}
   \item $\ourSystem$-GPT-4o mentions: "The exploitation of Arctic resources raises environmental concerns and potential conflicts over territorial claims. Increased maritime activity in the Arctic also brings risks of oil spills, noise pollution affecting marine life, and the introduction of invasive species."
   \item $\ourSystem$-Claude states: "These new economic activities come with significant challenges and potential conflicts. The competition for resources in the Arctic is intensifying, leading to increased militarization and the need for new governance structures to manage conflicting claims."
   \end{itemize}
   In contrast, Temporal-GPT-4o and Temporal-Claude primarily focus on the opportunities without thoroughly addressing associated challenges.
\end{enumerate}


\begin{table}[!h]
\centering
\begin{tabular}{|l|c|c|}
\hline
\textbf{Model} & \textbf{Retrieval Performance (\%)} & \textbf{Hypothesis Quality (\%)} \\
\hline
Temporal-Claude-Sonnet-3.5 & 45.46 & 26.88 \\
Temporal-GPT 4o & 35.83 & 18.28 \\
$\ourSystem$-Claude-Sonnet-3.5 & \textbf{63.54} & \textbf{34.68} \\
$\ourSystem$-GPT 4o & 57.33 & 34.30 \\
\hline
\end{tabular}
\caption{Performance Comparison: Reasoning Algorithms and LLM Combinations.}
\label{tab:performance-comparison}
\end{table}

\section{Conclusions}

This paper advances AI reasoning by introducing lateral thinking for agents to address low-specificity anticipatory and causal questions in streaming setting. We present a framework for generating lateral thinking queries and datasets, and introduce $\ourSystem$, a multi-agent system implementing System-2 reasoning. Preliminary studies reveal $\ourSystem$'s superior performance over single-agent systems, suggesting a novel phenomenon: lateral information flow across long-distance agent interactions, combined with fine-grained belief management, yields richer information contexts and enhanced reasoning. Rigorous testing of this hypothesis on large-scale datasets will be subject of future work.

\section{Acknowledgments}
This work was supported by the Center for AI and Center for Cloud Computing at Pacific Northwest National Laboratory, a multi-program national laboratory operated by Battelle for the U.S. Department of Energy.
\bibliography{citations}
\bibliographystyle{ieeenat_fullname}

\newpage
\appendix
\section{Details of $\ourSystem$ Architecture}

\subsection{Agent network initialization}
\label{subsec:Agent network initialization}

\begin{mdframed}

\textbf{User}: I want to know as soon as a geopolitical event between China and Taiwan that will affect the supply chain for American semiconductor companies happens.

\textbf{Prompt}: Given the following query construct a numbered list of 4 areas of research that should be focussed on to answer the query.

\textbf{Master Agent}: To address the query effectively, the following areas of research should be focused on:

1. **Geopolitical Tensions and Policies**:
\begin{itemize}
    \item Monitoring diplomatic relations and potential conflicts between China and Taiwan.
    \item Analysis of government policies, sanctions, and trade agreements affecting both nations.
    \item Updates on military activities or significant political statements that could indicate rising tensions.
\end{itemize}

2. **Supply Chain Dynamics**:
\begin{itemize}
   \item Identification of critical semiconductor components and their origins in China and Taiwan.
   \item Tracking of existing and potential supply chain disruptions within the semiconductor industry.
   \item Assessing dependency of American semiconductor companies on Chinese and Taiwanese suppliers.
\end{itemize}

3. **International Trade and Economics**:
\begin{itemize}
   \item Analysis of trade flows and economic interdependencies between China, Taiwan, and the United States.
   \item Monitoring changes in import/export regulations that could affect semiconductor industries.
   \item Studying economic indicators that might predict or result from geopolitical disruptions.
\end{itemize}

4. **Industry News and Developments**:
\begin{itemize}
   \item Real-time tracking of news within the semiconductor industry, focusing on announcements from key companies.
   \item Updates on corporate strategies and responses to geopolitical events by American semiconductor companies.
   \item Following industry analyses and expert opinions on potential impacts of geopolitical changes.
\end{itemize}

By focusing research on these areas, one can not only anticipate but also respond effectively to any geopolitical event between China and Taiwan that might affect the supply chain for American semiconductor companies.
\end{mdframed}

\begin{figure}[htb]\centering
    {\includegraphics[width=0.9\textwidth]{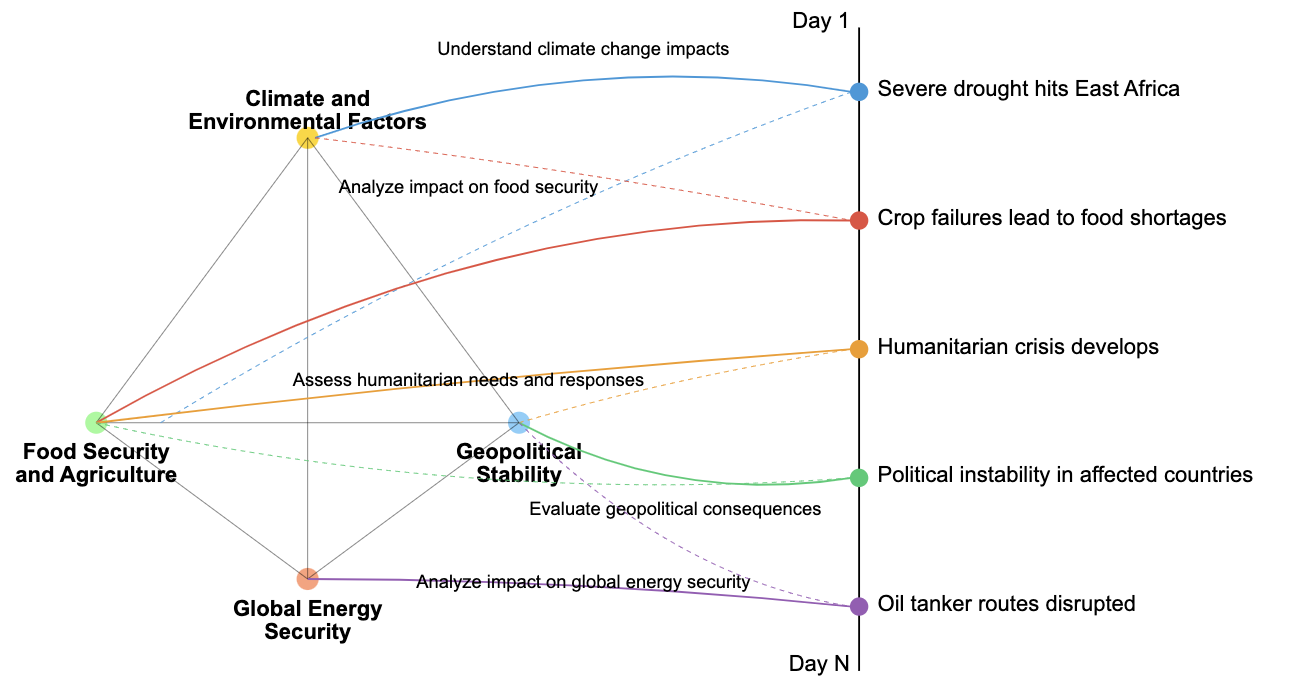}}
\caption{Illustration of a scenario where a drought in one part of the world can affect the energy security in distant places.}
\label{fig:salt_africa}
\end{figure}

\begin{figure}[htb]\centering
    {\includegraphics[width=0.9\textwidth]{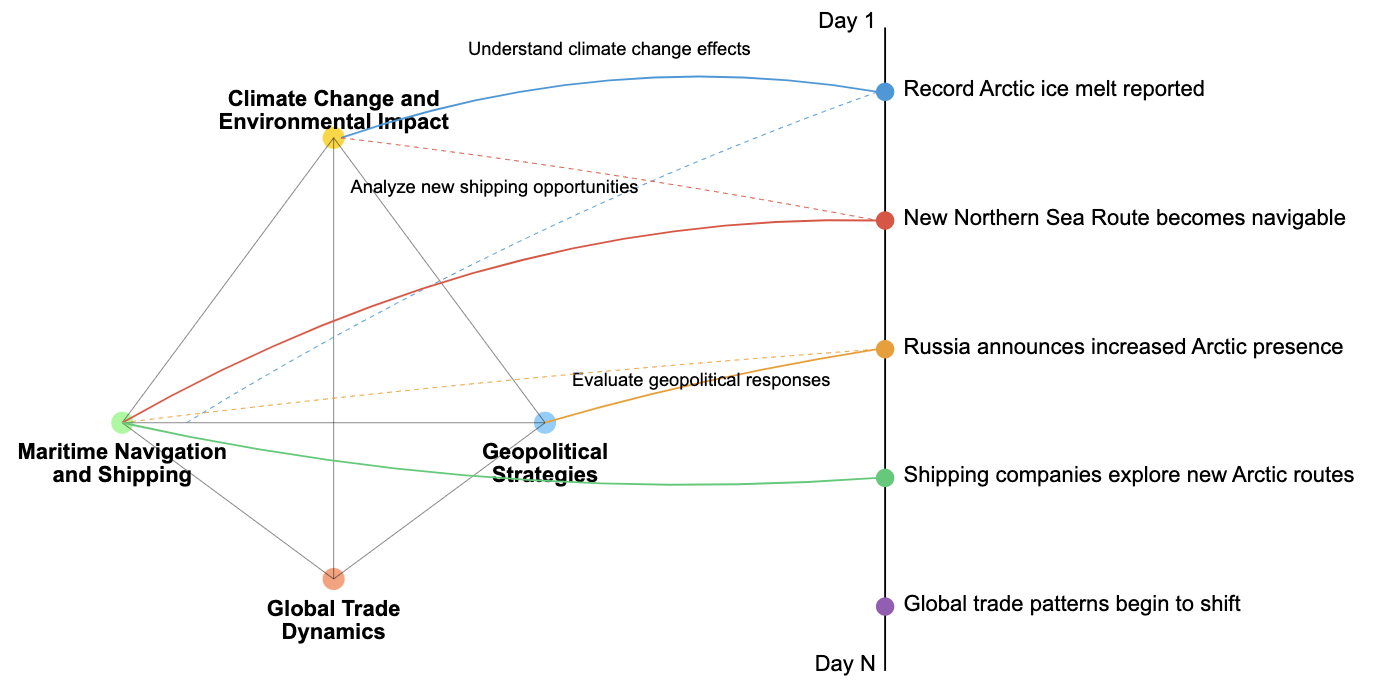}}
\caption{Illustration of a scenario where a climate change induced event can lead to upheaval in an industry.}
\label{fig:salt_arctic}
\end{figure}

\newpage

\section{Benchmarking}
\label{sec:Baseline Implementations}
\subsection{Implementation}
The SingleTemporalReasoningAgent processes streaming information and generates evolving hypotheses over time. Its core functionality involves maintaining a temporal context of past assessments while continuously integrating new information. The agent processes articles in chronological batches, capturing the progression of events and insights. For each user query, it generates hypotheses by considering both the current batch of articles and, optionally, the previous hypothesis, implementing a form of persistence in its reasoning process. This method enables the agent to build upon past knowledge, refine its assessments, and provide updated insights as new information becomes available. The agent employs a planner (or master agent) for hypothesis generation, adding adaptability to the reasoning process. As detailed in Algorithm 1, the agent manages the flow of information by sorting and filtering articles, generating hypotheses, and presenting results, creating a systematic process for continuous reasoning over time-based data streams.

\begin{algorithm}[H]
\SetAlgoLined
\KwIn{User queries $Q$, Planner $P$, Articles $A$, Parameters $params$}
\KwOut{Temporal context of hypotheses $H$}

Initialize $H$ as empty dictionary\;
Sort $A$ chronologically\;
$A \gets \text{FilterArticlesByDateRange}(A, params.start\_date, params.end\_date)$\;

\While{$A$ is not empty}{
    $batch \gets \text{GetNextBatch}(A)$\;
    
    \ForEach{query $q$ in $Q$}{
        $prompt \gets \text{GeneratePrompt}(batch, q, H[q].\text{last})$\;
        $hypothesis \gets P.\text{GenerateHypothesis}(prompt)$\;
        $H[q].\text{append}(hypothesis)$\;
        $\text{PresentHypothesis}(hypothesis)$\;
    }
    
    $\text{WaitForNewData}()$\;
}

\caption{SingleTemporalReasoningAgent}
\end{algorithm}

The algorithm maintains a temporal context of hypotheses for each user query, updating them as new batches of articles are processed. This allows the agent to incorporate past information while generating new hypotheses, resulting in a continuous stream of updated assessments based on the incoming information.

\newpage
\subsection{Query Configurations}

\begin{figure}[!h]
    \centering
    \includegraphics[width=0.85\textwidth]{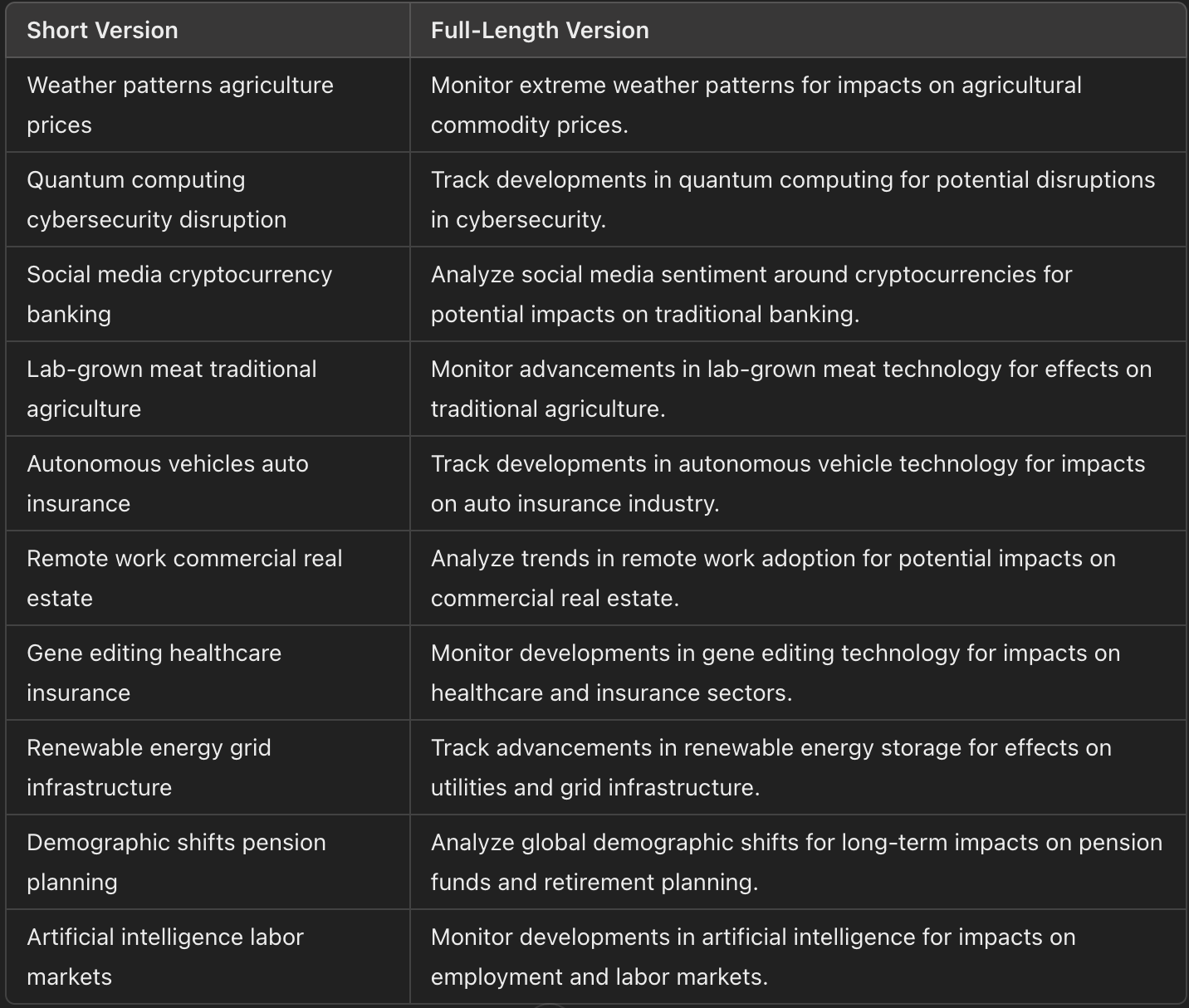}
    \caption{Example lateral thinking queries.}
    \label{fig:lt_queries}
\end{figure}


\begin{figure}[h]
    \centering
    \includegraphics[width=0.8\textwidth]{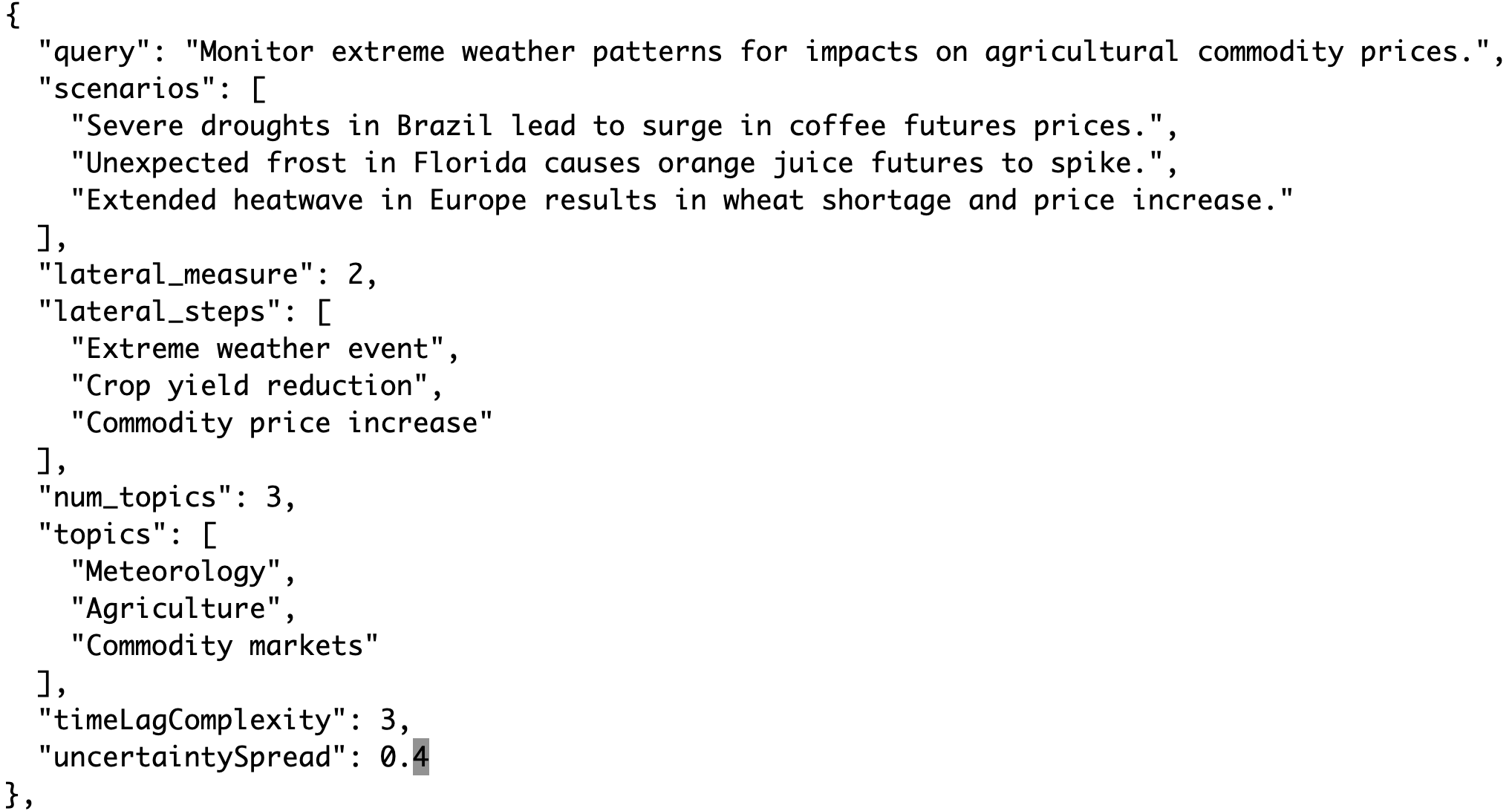}
    \caption{Example query-scenario pairs with corresponding metrics.  Note that the metrics are obtained on the basis of the ``query" field.}
    \label{fig:lt_query_metrics}
\end{figure}

\end{document}